\documentclass[10pt,twocolumn,letterpaper]{article}

\usepackage{iccv}      

\definecolor{iccvblue}{rgb}{0.21,0.49,0.74}
\usepackage[pagebackref,breaklinks,colorlinks,allcolors=iccvblue]{hyperref}
\usepackage{tikz}
\usepackage{amsmath}
\usepackage{dsfont}  
\usepackage{pgfplots}
\pgfplotsset{compat=1.17}
\usepackage{pifont}
\usepackage{xcolor}
\usepackage{fontawesome5}

\newcommand{\xmark}{\ding{55}}%
\newcommand{\model}{MATR}
\newcommand{\data}{SportsMoments}
\newcommand{\actdata}{ActivityNet-VRL}

\usepackage{enumitem}
\usepackage{algpseudocode}
\usepackage[font=small,labelfont=bf]{caption}
\usepackage{array}
\usepackage{multirow}
\usepackage{booktabs}
\usepackage{algorithm}
\usepackage{xparse}

\usepackage{bm}
\usepackage{threeparttable}
\usepackage{etoolbox}

\usepackage{listings}
\usepackage{mwe} %
\usepackage{makecell}
\usepackage{tabularx}
\usepackage{pifont}%
\usepackage[accsupp]{axessibility}

\definecolor{ImageDark}{rgb}{0,0.3,0.8}
\definecolor{VideoDark}{rgb}{.5,.0,.5}
\definecolor{DepthDark}{rgb}{0,.5,0}
\definecolor{AudioDark}{rgb}{0.11764705882352941, 0.5647058823529412, 1.0}
\definecolor{ThermalDark}{rgb}{0.8823529411,0.63725490196,0.0156862745}
\definecolor{IMUDark}{rgb}{0.6235294117647059, 0.27058823529411763, 0.4627450980392157}

\def\paperID{7754} 
\def\confName{ICCV}
\def\confYear{2025}

\title{Aligning Moments in Time using Video Queries}

\author{
Yogesh Kumar\textsuperscript{1*} \quad
Uday Agarwal\textsuperscript{1*} \quad
Manish Gupta\textsuperscript{2} \quad
Anand Mishra\textsuperscript{1} \\
\textsuperscript{1}Indian Institute of Technology Jodhpur \quad
\textsuperscript{2}Microsoft \\
{\tt\small \{kumar.204, agarwaluday, mishra\}@iitj.ac.in, gmanish@microsoft.com}
}

\begin{document}

	\urlstyle{same}

\maketitle

\begingroup
\renewcommand\thefootnote{*}
\footnotetext{Equal Contribution}
\endgroup

\begin{abstract}

Video-to-video moment retrieval (\emph{Vid2VidMR}) is the task of localizing unseen events or moments in a target video using a query video. This task poses several challenges, such as the need for semantic frame-level alignment and modeling complex dependencies between query and target videos.
To tackle this challenging problem, we introduce \model{} (\underline{M}oment \underline{A}lignment \underline{TR}ansformer), a transformer-based model designed to capture semantic context as well as the temporal details necessary for precise moment localization. \model{} conditions target video representations on query video features using dual-stage sequence alignment that encodes the required correlations and dependencies. 
These representations are then used to guide foreground/background classification and boundary prediction heads, enabling the model to accurately identify moments in the target video that semantically match with the query video. Additionally, to provide a strong task-specific initialization for \model{}, we propose a self-supervised pre-training technique that involves training the model to localize random clips within videos. 
Extensive experiments demonstrate that \model{} achieves notable performance improvements of 13.1\% in R@1 and 8.1\% in mIoU on an absolute scale compared to state-of-the-art methods on the popular \actdata{} dataset. Additionally, on our newly proposed dataset, \data{}, \model{} shows a 14.7\% gain in R@1 and a 14.4\% gain in mIoU on an absolute scale over strong baselines. We make the dataset and code public at: \href{https://github.com/vl2g/MATR}{https://github.com/vl2g/MATR}.
\end{abstract}
\section{Introduction}\label{sec:intro}

\begin{figure}[!t]
    \centering
      \includegraphics[width=0.48\textwidth]{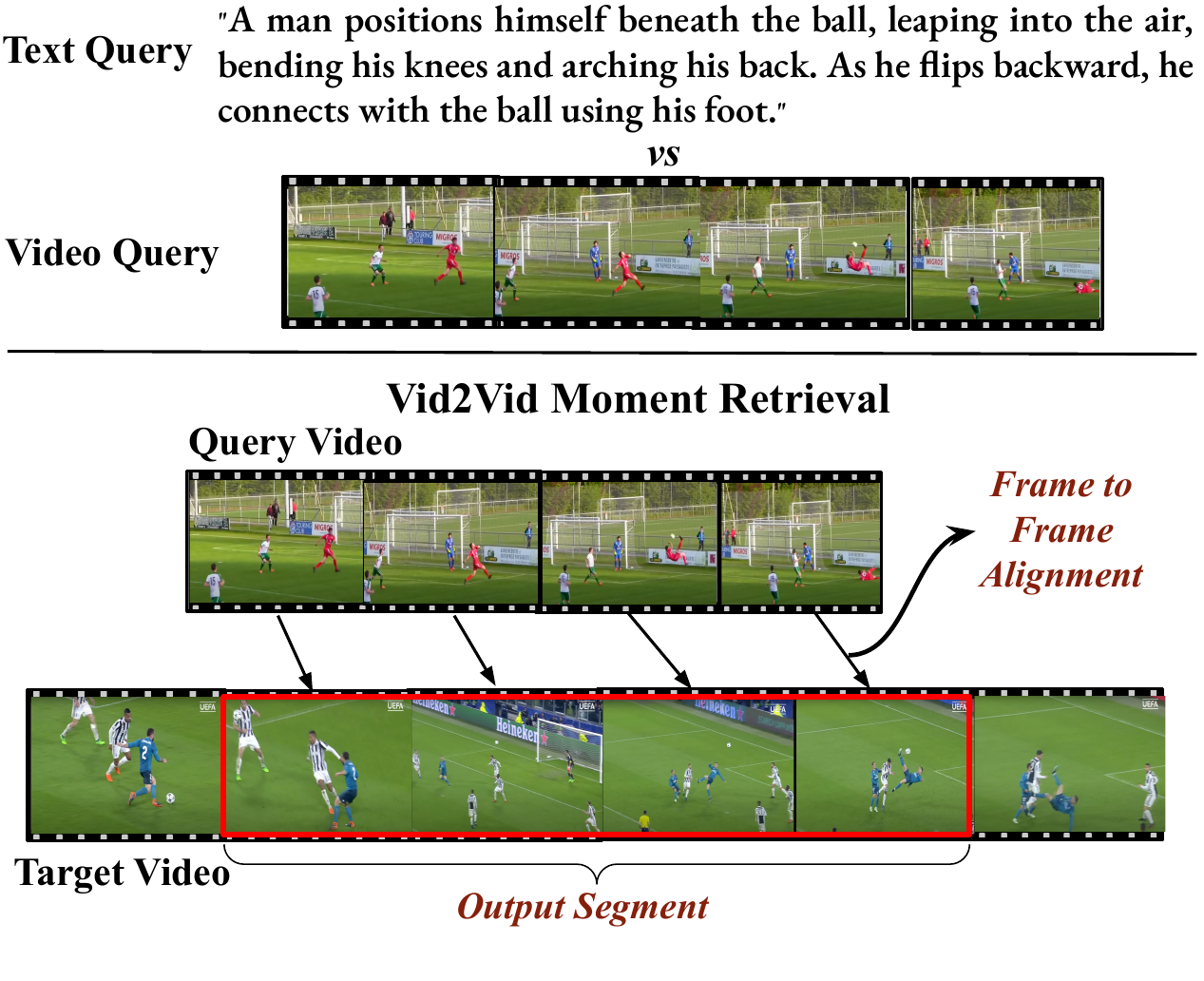}
    \caption{\label{fig:goal} (\textbf{Top}): Activity where text falls short to explain complex action or events, such as \textit{bicycle kick}, necessitating the need for a more intuitive query modality for localizing semantically matching moments. (\textbf{Bottom}): 
 Shows the Vid2Vid moment retrieval setting, which is the goal of this work.}
    
\end{figure}
Video moment retrieval is the task of temporally localizing the start and end times of a moment\footnote{A moment is a continuous segment of frames within a target video that best represents the actions, events, or interactions described by the query.} in a target video described by a given query. Although text-based video moment retrieval has been extensively explored~\cite{moment_detr,QD-DETR,cg-detr,univtg,Taskweave}, it often poses challenges for users attempting to describe specific moments verbally. For example, consider a spectacular \emph{bicycle kick} in soccer (Fig.~\ref{fig:goal} top). Although an expert can search by naming or describing it in detail, a beginner in soccer may struggle to describe this impressive move accurately. They might say something like ``The player kicked the ball while in the air'', which lacks the nuance needed to convey the athleticism and artistry involved. Further, such a description may lead to poor results for someone searching for similar moments. In contrast, if a soccer trainee shows a short video clip of a \textit{bicycle kick}, the input is rich and crisp, making it easier for a retriever to locate similar moments in a target soccer match video. This approach aligns with how users naturally would prefer to search content within the same (video) modality, 
making it a more effective method for retrieving specific moments. Fig.~\ref{fig:goal} (bottom) shows a video \textit{bicycle kick} query in soccer that is localized in a target video. 

Therefore, in this work, we study video-to-video moment retrieval (\emph{Vid2VidMR}) where the aim is to temporally localize a moment in the target video with a high semantic match between the moment and the query video. \emph{Vid2VidMR}, a task formally introduced by Feng et al.~\cite{video_reloc}, has many potential application areas, such as sports video analytics, educational content creation and e-learning, and surveillance systems. 
This task is challenging and requires semantic frame-level alignment and modeling complex dependencies between query and target videos. 
The need for a semantic understanding of video content and variety in video length, context, and action speed calls for adaptive models capable of generalizing across diverse scenarios. Addressing these challenges demands advanced temporal video modeling techniques.

To address the aforementioned challenges, we propose \model{} 
 (\underline{M}oment \underline{A}lignment \underline{TR}ansformer) -- a method that uses explicit `dual-stage sequence alignment' to capture the required correlation and temporal details essential for accurate moment localization. By conditioning target video representations on query features, \model{} produces query-aligned representations that encode the necessary correlations and temporal dependencies between the two videos. We use differentiable dynamic time warping loss~\cite{soft-dtw} for aligning the query and target videos, and represent the target video by conditioning it on the query video to focus on correlated temporal features. These representations guide a classification head to discriminate relevant moments from the irrelevant background and a boundary prediction head to mark the start and end of the identified moment in the target video. Further, to enhance our model's generalization using unlabeled videos, we introduce a self-supervised pre-training strategy which involves training \model{} to localize randomly sampled clips within the same video, enabling it to learn the moment localization skill in a self-supervised manner. 
 
We evaluate \model{} on public \actdata{}~\cite{video_reloc} benchmark and on our newly introduced dataset on sports domain, viz. \data{}, covering two of the most popular sports, namely soccer and cricket.
Our approach achieves significant performance gains, with an improvement of 13.1\% and 8.1\% in R@1 and mIoU, respectively, on \actdata{} outperforming the state-of-the-art methods. Furthermore, \model{} outperforms the implemented strong baselines on our proposed dataset with 14.7\% gains in R@1 and 14.4\%  gains in mIoU, all on an absolute scale. 

Our contributions are as follows: (i) We introduce \model{}, which uses explicit dual-stage sequence alignment within a transformer framework between target and query video to capture temporal correlations and dependencies for accurate moment localization. (ii) We propose a self-supervised pre-training objective that enhances model initialization by understanding rich video structure without requiring any labeled data. (iii) We conduct extensive experiments and ablations to study the efficacy of our framework against competitive baselines and state-of-the-art methods. Our findings offer valuable insights into our design choices, and our approach advances the state-of-the-art on \emph{Vid2VidMR}.

\section{Related Work}
\label{s:rel_work}
\noindent\textbf{Video Moment Retrieval (VMR):} VMR has recently gained significant interest in the research community~\cite{DiDeMO, TVR, CTRL, moment_detr, QD-DETR,univtg,Taskweave, cg-detr, uvcom, UnLOC, UMT,mun2020LGI, chapvidmr}. Unlike video action understanding tasks such as action classification~\cite{TwoStreamNets, TSM, TPN,C3D, Kinetics, S3D, SlowFast,TQN} or temporal action localization~\cite{tal2019graph,talenriching,tal}, VMR focuses on identifying segments that semantically align with a broader range of queries, which may describe complex and context-specific moments. Based on the query modality, existing VMR methods can be broadly grouped into (i) textual query-based approaches like Moment-DETR~\cite{moment_detr}, QD-DETR~\cite{QD-DETR}, Uni-VTG~\cite{univtg}, and (ii) video query-based approaches like GDP~\cite{GDP}, FFI+SRM~\cite{FFI-SRM}, SRL~\cite{SRL}. Our work falls under video query-based VMR (\emph{Vid2VidMR}). However, in addition to developing a new approach tailored for video query, we also adapt several text-query-based methods to make them suitable for \emph{Vid2VidMR} and compare our method against both text and video query-based methods. 

\noindent\textbf{Alignment in Videos:} Alignment has been a thrust area in the video understanding community. It has been applied to a wide range of video understanding tasks, including video retrieval~\cite{al_wang2021t2vlad},  procedural steps alignment~\cite{DropDTW}, action recognition~\cite{al_dogan2018neural}, anomaly detection~\cite{TCC}, movie understanding~\cite{Argaw2023LongrangeMP} and video synchronization~\cite{al_arao2003view,al_padua2008linear}. Researchers have also explored sequence alignment for the text-VMR task, e.g., Mithun et al.~\cite{mithun2019weakly} used sequence alignment between CNN representation of frames and GRU representation of text query to perform text-VMR. Jung et al.~\cite{Jung_2025_WACV} used alignment to enhance semantic understanding between query and target video at the abstract level for the text-VMR task. However, their alignment is not at the sequence level. 

Compared to the existing literature, \model{} goes beyond traditional sequence alignment by introducing a dual-stage alignment mechanism, leveraging transformer-based feature fusion, and maintaining flexibility with multiple alignment strategies. These innovations enable it to achieve more accurate and context-aware video moment retrieval compared to existing approaches.

\begin{figure*}[t!]
    \centering
      \includegraphics[width=\textwidth]{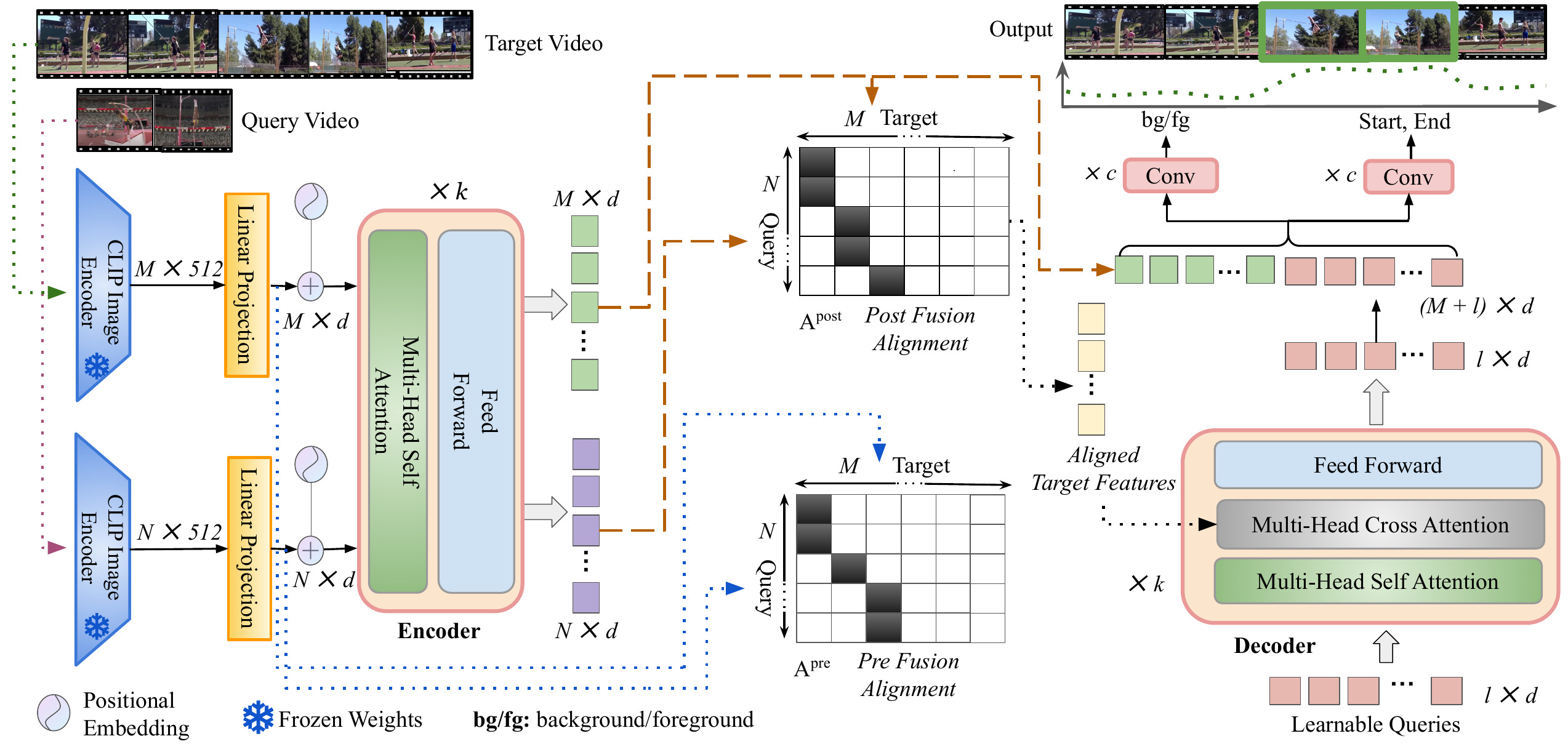}
  \caption{\textbf{\underline{M}oment \underline{A}lignment \underline{TR}ansformer (\model{})} Architecture. We represent the target video as a query-aligned representation capturing the required correlation and dependencies with query video at the abstract level (output of encoder) and finer level (output of decoder). These features are used to extract foreground target frames, as well as localize the relevant moment in terms of start and end times via two prediction heads. The alignment is computed using representations both before and after the encoder (shown using blue dotted and red dashed lines, respectively). \textbf{[Best viewed in color]}.}

  \label{fig:model}
  \end{figure*}

\section{\textbf{The \model{} Model}}
\label{sec:method}
Our objective is to temporally localize a moment in a target video $V_t$ using a query video $V_q$. We represent the target and query videos as sequences of $M$ and $N$ uniformly sampled frames, respectively. We refer to this problem as \emph{Vid2VidMR}. In this work, we present \underline{M}oment \underline{A}lignment \underline{TR}ansformer (MATR) -- a transformer-based architecture that leverages explicit alignment for precise moment localization. MATR transforms the target video into a query-aligned representation via a transformer framework to exploit finer-level correlation and dependencies for precise moment localization. 

\subsection{\textbf{Motivation behind MATR Architecture}}
Video moments vary significantly in duration, motion patterns, and visual appearance, making their retrieval inherently challenging. Additionally, preserving the correct temporal order of events is crucial for accurate localization, as moments often involve complex interactions that unfold over time. A robust \emph{Vid2VidMR} model must, therefore, capture both high-level semantic relationships and fine-grained frame dependencies to ensure precise alignment between the query and target video.

To address these challenges, \model{} incorporates an explicit \emph{dual-stage sequence alignment} strategy within a transformer-based encoder-decoder framework. This strategy enables \model{} to learn a \emph{query-aligned representation} of the target video by combining the abstract representation from the encoder with the refined fine-grained features from the decoder. In doing so, it facilitates precise moment localization by leveraging both global semantic alignment and fine-grained temporal dependencies between the query and target videos. Figure~\ref{fig:model} illustrates the overall architecture of \model{}, which we discuss in detail in the following.

\subsection{\textbf{Architecture Details}}

\noindent{\textbf{Input Representation.}} 
We uniformly sample frames every 2 seconds to obtain $M$ and $N$ frames from the target and query videos, respectively. We encode these frames using a frozen CLIP~\cite{clip} (ViT-B/32) encoder along with a linear projection module. CLIP is applied separately to both videos to obtain target video embeddings $\mathrm{E}_t \in \mathbb{R}^{M \times d}$ and query video embeddings $\mathrm{E}_q \in \mathbb{R}^{N \times d}$. The linear projection module is a two-layer perceptron, each with layer normalization and dropout. It maps $512$-dimensional CLIP embeddings to $d$-dimensional outputs. The resulting projected features ($\mathrm{E}_t$ and $\mathrm{E}_q$) are concatenated along the sequence length dimension to form the input to the transformer encoder, denoted as $\mathrm{E}_c=[\mathrm{E}_t;\mathrm{E}_q] \in \mathbb{R}^{(M + N) \times d}$.

\noindent{\textbf{Encoder.}}
The combined target and query video representation ($E_c$) is processed through a series of $k$ standard Transformer~\cite{Transformer} encoder layers, each comprising a multi-head self-attention mechanism and a feed-forward network. Further, following the prior works~\cite{DETR, Img_Transformer, Attn_ConvNet}, fixed positional encodings are added to the input of each attention layer to preserve temporal order. The encoder generates a fused representation $[\mathrm{E_t^g}; \mathrm{E_q^g}]$ of the target video conditioned on the query video. Here $\mathrm{E_t^g}\in \mathbb{R}^{M \times d}$ and $\mathrm{E_q^g}\in \mathbb{R}^{N \times d}$ represent the target and query parts respectively.

\noindent{\textbf{Dual-stage Sequence Alignment}}. 
The transformer encoder computes effective features by performing a joint understanding (or fusion) of target and query video frames. The final goal of \emph{Vid2VidMR} is to localize a moment in the target video that semantically matches the content of the query video. In other words, we would prefer the features of the moment in the target to align strongly with the features of the query video. Therefore, we perform dual-stage sequence alignment, i.e, before and after the encoder. For alignement, we choose soft-DTW~\cite{soft-dtw}\footnote{Soft-DTW allows non-linear alignments between sequences of different lengths, is differentiable, robust to variations in speed, and can handle noise and outliers, though other alignment algorithms, such as TCC~\cite{TCC} and DropDTW~\cite{DropDTW} can also be used in our model.} and perform the alignment as follows. 

\noindent \underline{Pre-fusion alignment:} Before encoder fusion, given the target and query video feature sequences, $\mathrm{E_t} = [\bm{e}_1^t, \ldots, \bm{e}_M^t] \in \mathbb{R}^{M \times d}$ and $\mathrm{E_q} = [\bm{e}_1^q, \ldots, \bm{e}_N^q] \in \mathbb{R}^{N \times d}$, soft-DTW outputs a binary alignment matrix $\mathrm{A}^{\text{pre}}$ and an alignment cost matrix $\mathrm{C}^{\text{pre}}$. $\mathrm{A}^{\text{pre}}\in \{0,1\}^{M \times N}$ is an alignment matrix such that: 
\begin{equation}
    \nonumber\mathrm{A}^{\text{pre}}_{i,j} = 
    \begin{cases} 
      1 & \text{if } \bm{e}_i^t \text{ is matched to } \bm{e}_j^q, \\
      0 & \text{otherwise.}
    \end{cases}
\end{equation}
further, soft-DTW finds the optimal alignment by minimizing the pre-fusion alignment loss \( \mathcal{L}_{\text{align}}^{\text{pre}} \), defined as:

\begin{equation}
\nonumber\mathcal{L}_{\text{align}}^{\text{pre}}= \mathrm{soft\text{-}DTW}_\gamma(\mathrm{A}^{\text{pre}}_{i,j}, \mathrm{C}^{\text{pre}}_{i,j}),
\end{equation}
\noindent such that 
aligned frames in the target are contiguous. Here $\gamma$ is a smoothing factor for soft-min operator. The value of $\gamma$ in Soft-DTW is selected empirically to balance smoothness and alignment fidelity. Note that $\mathrm{C}^{\text{pre}}$ is the alignment cost matrix with elements defined by cosine similarity:
\begin{equation}
    \nonumber\mathrm{C}_{i,j} = 1 - \frac{\langle \bm{e}_i^t, \bm{e}_j^q \rangle}{\|\bm{e}_i^t\| \|\bm{e}_j^q\|}.
\end{equation}
Pre-fusion alignment enhances the semantic representation of target and query video. These enhanced representations are further processed by the encoder. The encoder outputs the fused representation of the target and query.  

\noindent\underline{Post-fusion alignment:} Given the post-fusion target and query video feature sequences, $\mathrm{E_t^g} \in \mathbb{R}^{M \times d}$ and $\mathrm{E_q^g} \in \mathbb{R}^{N \times d}$, soft-DTW outputs a binary alignment matrix $\mathrm{A}^{\text{post}}$ and an alignment cost matrix $\mathrm{C}^{\text{post}}$, soft-DTW achieves optimal alignment between $\mathrm{E_t^g}$ and $\mathrm{E_q^g}$ by minimizing the post-fusion alignment loss \( \mathcal{L}_{\text{align}}^{\text{post}} \), defined as:

\begin{equation}
\nonumber\mathcal{L}_{\text{align}}^{\text{post}}= \mathrm{soft\text{-}DTW}_\gamma(\mathrm{A}^{\text{post}}_{i,j}, \mathrm{C}^{\text{post}}_{i,j}).
     \label{eq:optimal_alignment_pre}
\end{equation}
Post-fusion alignment refines target video features by leveraging fused query-target representations, ensuring fine-grained semantic matching for precise moment localization.

\noindent\textbf{Decoder}. The decoder in \model{} further refines the target video representation by processing aligned query-target features, enabling fine-grained temporal matching and enhancing frame-level precision for accurate moment localization. To this end, the post-fusion alignment matrix $\mathrm{A}^{\text{post}}$ defines a contiguous sub-sequence $\mathrm{E_t^g}[s:e]$ which aligns best with the query video. These aligned target features, capturing query-aligned fine-grained information, are passed as input to the decoder for further refinement.
The decoder consists of $k$ Transformer layers, each containing a multi-head self-attention layer, a multi-head cross-attention layer, and a feed-forward network. The input to the decoder comprises fixed size $l$ learnable query vectors denoted by $\mathrm{Q} \in [q_1,\ldots, q_{l}] \in \mathbb{R}^{l \times d}$. These queries guide the extraction of refined features relevant to moment localization. As these queries are processed through each decoder layer, they are refined to capture intricate temporal dependencies, with positional encodings applied at each attention layer. The cross-attention layers enable interaction between $\mathrm{E_t^g}[s:e]$ from the encoder (which serve as keys and values) and representations of moment queries (which serve as queries). The final decoder output, $\mathrm{E_t^l} \in \mathbb{R}^{l \times d}$, provides a refined fine-grained set of target video features conditioned on the query video. 

Finally, the encoder and decoder representations are combined as $\mathrm{E_f} = [\mathrm{E_t^g}; \mathrm{E_t^l}]$, making the final target video representation query-aligned. The final \emph{query-aligned representation}, combining the abstract and fine-grained semantics, serves as input for the prediction heads.

\noindent{\textbf{Prediction Heads.}} 
The prediction heads consist of three ($c = 3$) sequential convolutional layers each having $d$, \( 1\times3 \) kernels followed by a ReLU activation. Finally, a sigmoid activation function is applied to produce the foreground predictions \( \hat{f}_i \). The model is trained using the following binary cross-entropy loss $\mathcal{L}_{fg}$ to distinguish foreground (relevant moment) from background predictions:
\begin{equation}
\nonumber\mathcal{L}_{fg} = -\left( f_i \log \hat{f}_i + (1 - f_i) \log (1 - \hat{f}_i) \right),
\end{equation}
where \( f_i \) is the ground truth label, with \( f_i = 1 \) and \( f_i = 0 \) indicating foreground and background, respectively.

In addition to foreground classification, the boundary prediction head is designed to estimate the start and end boundaries for the moment at every position. This head shares the same initial structure as the classification head but differs in the final output, producing two channels for left and right boundary offsets relative to the current position. Specifically, given \( \mathrm{E_f} \), the boundary prediction head outputs predicted offsets \( \hat{d}_i = \{\hat{d}_i^L, \hat{d}_i^R\} \) for each position \( i \in [0,M-1]\), where \( \hat{d}_i^L \) and \( \hat{d}_i^R \) represent the left and right boundary offsets, respectively. The boundary prediction head is trained with a combination of smooth $L1$ and generalized intersection over union (IoU) loss~\cite{rezatofighi2019generalized}, specifically applied to foreground positions (\( f_i = 1 \)), defined as:%
\begin{equation}
\nonumber\mathcal{L}_{\text{seg}} = \mathds{1}_{f_i = 1} \left[ \lambda_{\text{L1}} \mathcal{L}_{\text{L1}} \left( \hat{d}_i, d_i \right) + \lambda_{\text{IoU}} \mathcal{L}_{\text{IoU}} \left( \hat{d}_i, d_i \right) \right].
\end{equation}
Here, \( d_i \) denotes the ground truth offset. The parameters \( \lambda_{\text{L1}} \) and \( \lambda_{\text{IoU}}\) weigh the contributions of the smooth L1 and IoU losses, ensuring that boundary localization focuses on the predictions identified as foreground. 

Convolutional layers operate along the temporal axis over the concatenated encoder query-conditioned and decoder-aligned target features. This preserves temporal continuity, enabling fine-grained predictions without boundary artifacts.

\noindent{\textbf{Overall Loss.}}  
The overall loss function \( \mathcal{L} \) is computed as a combination of multiple objectives, accounting for alignment costs (pre- and post-fusion alignment), foreground/background classification, and boundary localization. For $S$ training samples, this multi-objective overall loss is defined as:
{\small \begin{eqnarray}
\nonumber\mathcal{L} = \frac{1}{S} \sum_{i=1}^{S} \left( \lambda_{fg}\mathcal{L}_{fg} + \lambda_{\text{seg}}\mathcal{L}_{\text{seg}}  + \lambda_{\text{align}}^{\text{pre}}\mathcal{L}_{\text{align}}^{\text{pre}} + \lambda_{\text{align}}^{\text{post}}\mathcal{L}_{\text{align}}^{\text{post}} \right),
\end{eqnarray}}
\noindent where the \( \lambda \)s control the importance of each component.

\noindent\textbf{Inference.} 
During inference, given a target video \( V_t \) and a query video \( V_q \), we pass both through the model to generate foreground probabilities \( \{\hat{f}_i\}_{i=0}^{M-1} \) and boundary predictions \( \{\hat{d}_i\}_{i=0}^{M-1} \) using the two prediction heads. To handle the densely generated boundaries, while predicting \( \hat{d}_i \), we apply 1-dimensional non-maximal suppression with a threshold of 0.7, filtering out highly overlapping boundaries and producing the final set of predictions. We choose \( \hat{d}_i \) corresponding to position $i$ with highest \( \hat{f}_i \).

\begin{figure}[t!]
    \centering
    \includegraphics[width=\linewidth]{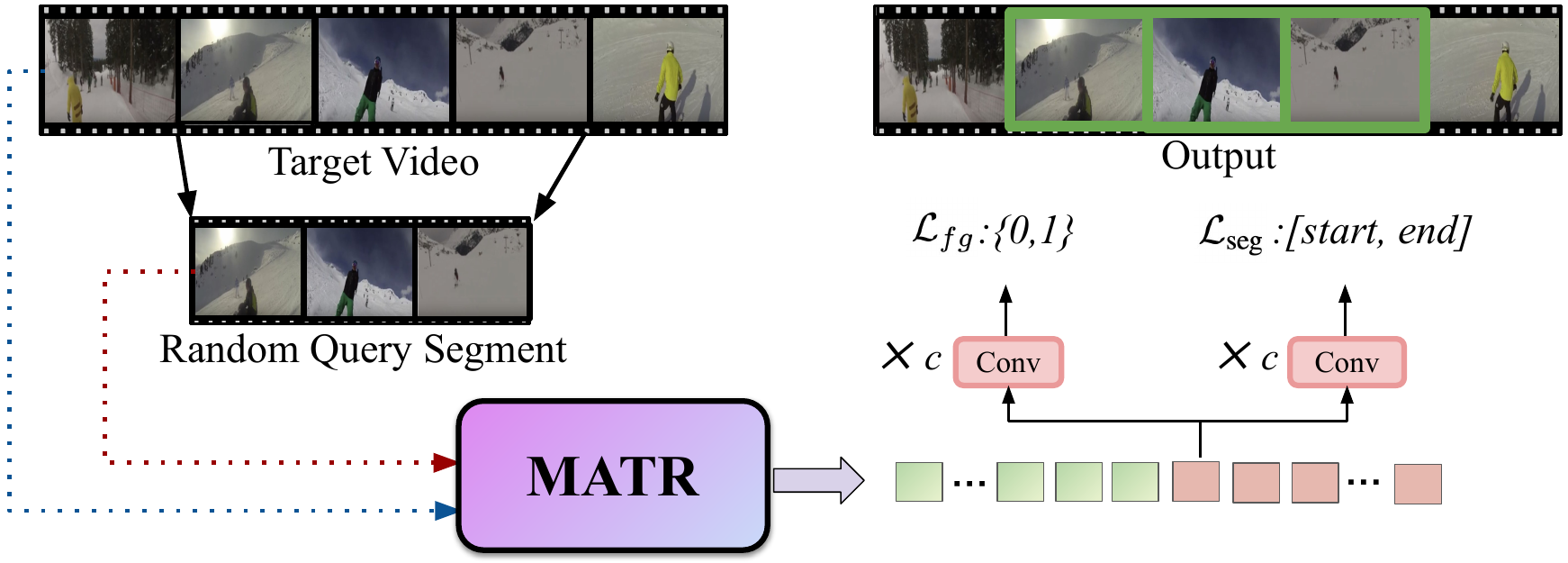}
    \caption{\textbf{Proposed self-supervised pre-training strategy:} Given a target video, a query clip is randomly sampled from it and processed by the model. The model then predicts the boundaries of the selected clip within the target video, highlighting the corresponding frames with a green border. \textbf{(Best viewed in color)}}
    \label{fig:pre_train}
\end{figure}

\subsection{\textbf{Self-supervised Pre-training Strategy}}\label{s:pt_obj}
The video representation learning community has leveraged self-supervised pre-training techniques designed for training effective encoders~\cite{vul1,vul2,vul3,vul4,vul5,vul6,vul7}, which can be finetuned for enhanced performance on downstream tasks. In this work, we introduce a self-supervised pre-training objective designed to improve temporal localization capabilities without relying on labeled data. Specifically, given a target video \( V_t \), a query clip \( V_q \) is randomly sampled from \( V_t \) as shown in Fig.~\ref{fig:pre_train}. The model is then trained to localize this query clip in \( V_t \). This pre-training objective closely aligns with the task of \emph{Vid2VidMR}. To enhance generalization capabilities and encourage the model to become robust to variations in both appearance and timing, we apply one of the following random augmentations to each of the query clips: reversing frames, adding gaussian noise, slowing down or speeding up the action, thereby doubling the number of pre-training samples. These augmentations introduce temporal and spatial variations, enabling the model to learn diverse representations.

The overall pre-training loss, \(\mathcal{L}_{pt}\), combines the foreground-background classification loss, the boundary prediction loss, pre-fusion and post-fusion alignment costs. The total loss is averaged over \( P \) pre-training samples as:
{\small\begin{eqnarray}
\nonumber\mathcal{L}_{pt} = \frac{1}{P} \sum_{i=1}^{P} \left( \lambda_{fg}^p\mathcal{L}_{\text{fg}} + \lambda_{seg}^p\mathcal{L}_{\text{seg}} + \lambda_{\text{align}}^{\text{pre}}\mathcal{L}_{\text{align}}^{\text{pre}} + \lambda_{\text{align}}^{\text{post}}\mathcal{L}_{\text{align}}^{\text{post}}\right),
\end{eqnarray}}
\noindent where the \( \lambda \)s control the importance of each component.

\section{Datasets} 
We use the following two datasets to compare our proposed approach with existing \emph{Vid2VidMR} methods and other strong baselines: 

\noindent{\textbf{(i) \actdata{} Dataset~\cite{video_reloc}:}}
\actdata{} is the popular benchmark dataset shared by Feng et al.~\cite{video_reloc}, based on the ActivityNet video understanding benchmark~\cite{ActivityNet} consisting of 200 action classes. The dataset is split into disjoint 160 classes for training and 20 classes each for the validation and test splits. Furthermore, the training set comprises of $\sim$ 463K query-target video pairs, while the validation set and the test set have 829 and 978 query-target video pairs, respectively. 

\noindent{\textbf{(ii) Our proposed \data{} Dataset\footnote{\url{https://github.com/vl2g/MATR/tree/main/sportsmoments}}:}}\label{s:data}
One of the most promising applications for \emph{Vid2VidMR} lies in the area of sports analytics. Although there exist large-scale sports datasets such as Sports-1M~\cite{Sports1M}, they are primarily tailored for broader sports video classification tasks and do not contain fine-grained sports actions and moments, such as ``Cover Drive,'' ``Ducking a Bouncer,'' ``Goal Kick,'' or ``Penalty.'' Secondly, although \actdata{} covers a wide range of action categories (including sports actions), it only encompasses broader categories such as ``Playing Polo'' and does not focus on capturing higher-granularity events within a sport. Towards filling this gap, we introduce the \data{} dataset, which consists of $\sim$770K query-target pairs annotated from 176.6 hours of complete match footage of two of the most popular sports, viz. soccer and cricket. We obtain a total of 80 full-length cricket and soccer full-length match videos from YouTube. We then curate a list of 29 action classes comprising 13 soccer and 16 cricket actions, respectively. Given this list, we employed two annotators, each with strong knowledge of these sports, to mark the start and end timestamps for the specified actions in the videos. 
We split \data{} into training, validation and test sets. The training split consists of approximately 750K pairs spanning 16 classes, while the validation and test splits each contain 10K pairs, covering four and nine classes, respectively. Action classes are disjoint across the train, validation, and test sets, ensuring no overlap. Additionally, each split includes cricket and soccer actions for a well-rounded distribution. 
Additionally, we leverage the unlabeled videos from the Kinetics700~\cite{Kinetics} dataset for pre-training.

\section{Experiments and Results}
\label{s:exps}

\subsection{Baselines}
\label{sec:base}
We experiment with an extensive set of strong baselines grouped into four categories as follows. 

\noindent(i) \textbf{Fully-supervised \emph{Vid2VidMR} methods:} We compare with \emph{Vid2VidMR} methods like CGBM~\cite{video_reloc}, GDP~\cite{GDP}, SRL~\cite{SRL}, FFI+SRM~\cite{FFI-SRM}, SST~\cite{SST} and Video-level match~\cite{video_reloc}. Further, Huo et al.~\cite{FFI-SRM} adapt text-VMR methods like VSLNet~\cite{VSLNet}, MABAN~\cite{MABAN} and 2D-TAN~\cite{2D-TAN} by replacing their text encoders with a C3D~\cite{C3D} feature extractor to extract query video features. All of these models have been trained on \actdata{}. 

\noindent(ii) \textbf{Vision-Language Models (VLMs):} 
VLMs like Video-LLaMa~\cite{videollama}, Video-LLaVa~\cite{videollava} and TimeChat~\cite{Ren2023TimeChat} have shown promising results for multimodal text-video tasks like Video VQA, captioning, etc. Unfortunately, they have not been pre-trained with video-video aligned data. First we represent the query video using a state-of-the-art captioner, i.e., mPLUG-OWL~\cite{mplugowl}. Next, the VLMs are zero-shot prompted to generate start and end moment timestamps given query caption and target video tokens.

\noindent(iii) \textbf{Text-VMR methods:} Moment-DETR~\cite{moment_detr}, QD-DETR~\cite{QD-DETR}, CG-DETR~\cite{cg-detr}, and UniVTG~\cite{univtg} have been originally proposed for text-VMR. We compare with five variants of these methods. The zero-shot variant (a) uses caption from mPLUG-OWL~\cite{mplugowl} to represent query video, and leverages the pretrained checkpoints. Variants (b) and (c) both use caption from mPLUG-OWL~\cite{mplugowl} to represent query video, but train a randomly initialized checkpoint (variant b) or finetune the pre-trained checkpoint (variant c), respectively. Variants (d) and (e) are equivalent to (b) and (c) where their CLIP text encoder is replaced by CLIP ViT/B-32 vision encoder, and therefore take query video directly as input along with the target video.

\noindent(iv) \textbf{Image-VMR methods:} Our work focuses on video moment retrieval using `video' queries. As a single image query may not effectively capture the temporal aspects of a video query, this comparison may not be appropriate. However, we still design baselines by representing the video by its key-frame (more precisely, the middle frame) using Text-VMR methods, namely Moment-DETR, QD-DETR and UniVTG.

\subsection{Implementation Details}
\label{s:imple-det}
We choose hidden dimension of $1024$ with $k=4$ layers in both encoders and apply a dropout rate of $0.1$ and  $0.5$ within the transformer and linear projection layers, respectively. Model weights are initialized using Xavier initialization~\cite{xavier}. To optimize the model parameters, we utilize AdamW optimizer~\cite{adam} with an initial learning rate of 1e-4 and a weight decay of 1e-4. Training was done on two NVIDIA A6000 GPUs. We trained our model for 200 epochs with a batch size of 1200 while using \actdata{}. For \data{}, we used 40 epochs with a batch size of 40. We set number of learnable queries $l=10$ and all $\lambda$s in both pre-training as well as finetuning losses to 1. 
We make our implementation and checkpoints available at: \href{https://github.com/vl2g/MATR}{https://github.com/vl2g/MATR}.

\begin{table}
\scriptsize
    \centering
    \setlength{\tabcolsep}{4pt}
    \begin{tabular}{ll lcccc}
    \toprule
       &\multicolumn{2}{c}{} & \multicolumn{2}{c}{\actdata{}} & \multicolumn{2}{c}{\data{}} \\
       \cmidrule(r){4-5} \cmidrule(r){6-7}
        &\multirow{1}{*}{Methods}   & Variants &  mIoU & R@1 & mIoU &  R@1 \\
        \midrule
        \multirow{10}{*}{\rotatebox{90}{Fully-supervised Methods}}&Random \cite{video_reloc} & - & 7.3 & 16.2 & - & -\\
        &Video Match \cite{video_reloc} & - & 12.4 & 24.3 & - & -\\
        &SST \cite{SST} & - & 17.1 & 33.2 & - & -\\
        &CGBM \cite{video_reloc} & - & 25.7 & 43.5 & - & -\\
        &GDP \cite{GDP} & - & 27.8 & \fbox{44.0} & - & -\\
        &SRL \cite{SRL} & - & 40.6 & 29.3 & - & -\\
        &2D-TAN \cite{2D-TAN} & - & 45.3 & 39.6 & - & -\\
        &MABAN \cite{MABAN} & - & 42.8 & 37.5 & - & -\\
       & VSLNet \cite{VSLNet} & - & 27.2 & 43.8 & - & -\\
        &FFI+SRM \cite{FFI-SRM} & - & \fbox{48.7} & 40.6 & - & -\\
        \midrule
        \multirow{4}{*}{\rotatebox{90}{VLMs}}&Video-LLaVA~\cite{videollava} & - & 15.1& 14.7 & 13.8 & 11.9\\
       & Video-LLaMA~\cite{videollama} & - & 14.7& 13.9 & 12.5 & 11.2\\
        &Video-LLaMA2~\cite{videollama2} & -& 17.6  & 15.2 & 14.7 & 13.4\\
       & TimeChat~\cite{Ren2023TimeChat}  & -& 26.4 & 23.8 & 22.6 & 21.3\\
        \midrule

        \multirow{2}{*}{\rotatebox{90}{I-VMR}}&Moment-DETR~\cite{moment_detr} & - & 35.6 & 32.5 & 25.2 & 18.5\\
       & QD-DETR~\cite{QD-DETR} & - & 37.2 & 38.1 & 27.7 & 20.7\\
        &UniVTG~\cite{univtg} & -& 38.8 & 42.0 & 34.6 & 37.2\\
        \midrule
        
\multirow{20}{*}{\rotatebox{90}{Text-VMR Methods}}&\multirow{6}{*}{Moment-DETR \cite{moment_detr}} & (a) ZS+T & 28.2 & 22.8 & 3.7 & 0.9\\
&& (b) Trained+T & 37.1 & 35.8 & 31.4 & 29.2\\
&& (c) Finetuned+T & 40.0 & 38.9 & 35.8 & 34.0\\
&& (d) Trained+V & 35.9 & 34.7 & 28.7 & 24.2\\
&& (e) Finetuned+V & 40.0 & 39.9 & 30.4 & 25.1\\
\cmidrule(r){2-7}
&\multirow{6}{*}{QD-DETR \cite{QD-DETR}} & (a) ZS+T & 25.4 &  22.2 & 6.3 & 2.6\\
&& (b) Trained+T & 38.4 & 39.0 & 29.8 & 27.2\\
&& (c) Finetuned+T & 41.6 & 45.2 & 36.0 & 33.5\\
&& (d) Trained+V & 39.7 & 41.0 & 27.2 & 25.1\\
&& (e) Finetuned+V & 42.5 & 42.7 & 35.4 & 30.6\\
\cmidrule(r){2-7}
&\multirow{6}{*}{UniVTG \cite{univtg}} & (a) ZS+T & 32.4 & 26.7 & 11.4 & 6.0\\
&& (b) Trained+T & 43.8 & 45.6 & 41.5 & 36.5 \\
&& (c) Finetuned+T & 45.8 & 46.4 & \underline{44.8} & 39.2 \\
&& (d) Trained+V & 48.4 & 49.8 & 43.2 & 39.5\\
&& (e) Finetuned+V & \underline{49.1} & \underline{50.7} & 43.6 & \underline{41.8} \\
\cmidrule(r){2-7}
&\multirow{6}{*}{CG-DETR \cite{cg-detr}} & (a) ZS+T & 25.9 & 21.9 & 3.3 & 1.6\\
&& (b) Trained+T & 39.4 & 38.4 & 35.1 & 31.3\\
&& (c) Finetuned+T & 41.0 & 43.2 & 35.4 & 32.5\\
&& (d) Trained+V & 40.0 & 41.7 & 36.6 & 35.1\\
&& (e) Finetuned+V & 40.1 & 41.7 & 37.2 & 34.8\\
        \midrule
\multirow{3}{*}{\rotatebox{90}{Ours}}&\multirow{3}{*}{\textbf{\model{}}} & Zero-shot & 32.6 & 30.1 & 31.8 & 30.7\\
& &Trained+V & 53.2 & 54.8 & 56.2 & 52.7\\
& & Finetuned+V & \textbf{56.8} & \textbf{57.1} & \textbf{59.2} & \textbf{56.5}\\
       \bottomrule 
    \end{tabular}
    \caption{Comparing \model{} with fully-supervised \emph{Vid2VidMR} methods, VLMs, image-VMR methods, and variants of Text-VMR methods. Trained implies random initialization. Finetuned implies initialization using a pre-trained (respective to each architecture) checkpoint. Query video can be represented using text (T) captions or as the video (V) itself. Since \model{} is inherently designed to take video query as input, experimenting with text (T) captions is not needed. For more details of baselines and their variants, please refer to Section~\ref{sec:base}. Results for the overall best, best among implemented baselines, and previously reported SOTA are highlighted in bold, underline, and box, respectively.} 
    \label{tab:main_table}
\end{table}

\subsection{Results and Discussion}

\noindent{\textbf{Main Results:}} Table~\ref{tab:main_table} shows our main results where we compare \model{} with fully-supervised \emph{Vid2VidMR} methods, VLMs, image-VMR methods, and five variants of each of the Text-VMR methods. As mentioned before, these variants correspond to zero-shot, finetuned, or trained setups. As usual, the trained setup implies random initialization, while the finetuned setup implies initialization using a pre-trained checkpoint (different for each architecture). Query video can be represented using text (T) captions obtained using mPLUG-OWL~\cite{mplugowl} or as video (V) itself. Since \model{} is inherently designed to take a video query as input, experimenting with text (T) captions and image query is not needed. Following previous work, we report the standard mean Intersection over Union (mIoU) and Recall@1 IoU=0.5 metrics.

\noindent{\textbf{Comparison with fully-supervised \emph{Vid2VidMR} methods:}} For a fair comparison, we directly use the reported results from original papers for fully-supervised methods on \actdata{}. Consequently, we do not report their results on \data{}. As shown in Table~\ref{tab:main_table}, \model{} outperforms the best method, FFI+SRM, with an 8.1\% gain in mIoU and achieves a 13.1\% improvement in R@1 over GDP, which had the highest recall. These results underscore \model{}'s superior ability to accurately localize moments using video queries. 

\noindent{\textbf{Comparison with VLMs:}} 
Table~\ref{tab:main_table} shows that our method \model{} outperforms all four VLM-based baselines by a significant margin. In comparison, the best-performing baseline, TimeChat~\cite{Ren2023TimeChat}, achieves 26.4 mIoU and 23.8 R@1 on \actdata{}, and 22.6 mIoU and 21.3 R@1 on \data{}. Other methods, such as Video-LLaVA~\cite{videollava}, Video-LLaMA~\cite{videollama}, and Video-LLaMA2~\cite{videollama2}, perform significantly worse. These results show that task-specific models for \emph{Vid2VidMR} are significantly better.
\\
\noindent{\textbf{Comparison with Text-VMR Methods:}}
Table~\ref{tab:main_table} shows that amongst all the \emph{five variants}, the finetuned variants are typically better than the trained variants, i.e., variant (c) is better than (b), and (e) is better than (d). On both \actdata{} and \data{}, UniVTG achieves the best results among variants that use video queries. However, our \model{} model outperforms UniVTG by 7.7\% in mIoU and 6.4\% in R@1 on \actdata{}. On \data{}, \model{} shows a further improvement, surpassing UniVTG by 15.6\% in mIoU and 14.7\% in R@1. These gains highlight the superior performance of \model{} compared to the text-VMR methods when used with video query input. When text captions are used to represent the query, we observe similar trends as in the video query input case. Among the baselines, UniVTG performs the best. However, \model{} outperforms it by 11\% in mIoU and 10.7\% in R@1 on \actdata{}. On \data{}, \model{} achieves 14.4\% higher mIoU and 17.3\% higher R@1 than UniVTG. These results demonstrate the superior performance of \model{} compared to the text-VMR methods when used with text query input for \emph{Vid2VidMR}.

\begin{table}[!t]
\scriptsize
    \centering
\begin{tabular}{c c c c c c}
\toprule
 \multicolumn{2}{l}{}  & \multicolumn{2}{c}{\actdata{}}  & \multicolumn{2}{c}{\data{}}\\
 \cmidrule(r){3-4} \cmidrule(r){5-6}
Pre-Fusion  & Post-Fusion & mIoU & R@1 & mIoU & R@1 \\ 
\hline
\xmark & \xmark  & 49.7 & 50.2 & 49.1& 46.6\\ 
\xmark & \checkmark & 52.3 & 52.9 & 53.1 & 48.2\\ 
\checkmark & \xmark & 50.4 & 51.2 & 52.4 & 46.7\\ 
\checkmark & \checkmark & \textbf{53.2} & \textbf{54.8} & \textbf{56.2} &\textbf{52.7} \\ 
\hline
\end{tabular}
\caption{Advantage of explicit dual-stage alignment of \model{}. For this study, experiments were performed without pre-training.}
\label{tab:abl_align}
\end{table}

\begin{table}[!t]
\scriptsize
    \centering
\begin{tabular}{c c c c c c}
\toprule
 \multicolumn{2}{l}{}  & \multicolumn{2}{c}{\actdata{}}  & \multicolumn{2}{c}{\data{}}\\
 \cmidrule(r){3-4} \cmidrule(r){5-6}
Pre-training & Augmentation & mIoU & R@1 & mIoU & R@1\\
 \midrule 
   \xmark & - & 53.2  &  54.8 & 56.2 & 52.7\\
  \checkmark & \xmark & 54.1  &  55.6 & 56.9 & 53.6\\
    \checkmark & \checkmark & \textbf{56.8} & \textbf{57.1} & \textbf{59.2} & \textbf{56.5}\\
 \hline
\end{tabular}
\caption{Effect of pre-training and augmentation in \model{}.}
\label{tab:pre-train}
\end{table}

 \begin{figure*}[!t]
    \centering
      \includegraphics[width=0.92\linewidth]{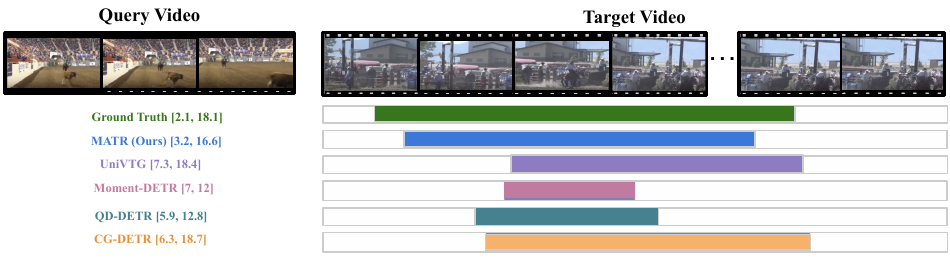}
  \caption{Visualization of \emph{Vid2VidMR} on a sample from \actdata{} for \emph{calf roping} action. Our proposed \model{} model shows improved generalization capabilities over the best-performing baseline methods (variant (e), i.e., finetuned+V). Start and end times for the ground truth and predicted moments are shown in brackets.}
  
  \label{fig:qual}
  \end{figure*}
  
\noindent{\textbf{Comparison with Image-VMR:}}
Table~\ref{tab:main_table} shows that the best-performing approach, UniVTG~\cite{univtg}, achieves a mIoU of 38.8 and 42.0 R@1 on ActivityNet-VRL and 34.6 mIoU and R@1 37.2  on \data{}. Although this is comparable to previous fully supervised methods, it still lags significantly behind our \model{}. This further goes to show the importance of temporal dynamics captured by video queries as compared to static image queries. 

\noindent{\textbf{Advantage of Dual Sequence Alignment:}}
Table~\ref{tab:abl_align} shows the impact of different alignment strategies in \model{} on both datasets. This study is done without any pre-training. We evaluate the effect of incorporating alignment at the pre-fusion and post-fusion stages. Our results indicate that the combined use of both pre-fusion and post-fusion alignment achieves the best performance, suggesting that alignment is important to ensure both before and after the encoder. Models with alignment either before (pre) or after (post) show moderate gains over neither. Removing post-fusion alignment hurts more than removing pre-fusion alignment, suggesting that post-fusion alignment is more important. Omitting alignment entirely results in the lowest scores. This highlights the importance of dual alignment for effectively capturing relevant temporal information for video query-based moment localization.

\noindent{\textbf{Effect of Pre-training:}} \label{s:pre}
We analyze the impact of pre-training and data augmentation on model performance in Table~\ref{tab:pre-train}. On \actdata{}, without pre-training, the model achieves 53.2 mIoU and 54.8 R@1. Adding pre-training without augmentation yields 54.1 mIoU and 55.6 R@1. The best results are obtained by using pre-training with augmentation, leading to a significant boost to 56.8 mIoU and 57.1 R@1. We got similar observations for the \data{} dataset. 

\begin{table}[!t]
\scriptsize
    \centering
\begin{tabular}{l cccc}
\toprule
 \multicolumn{1}{l}{}  & \multicolumn{2}{c}{\actdata{}}  & \multicolumn{2}{c}{\data{}}\\
 \cmidrule(r){2-3} \cmidrule(r){4-5}
 & mIoU & R@1 & mIoU & R@1\\
\midrule
Pre-fusion alignment ($\mathrm{A^{\text{pre}}}$) & 34.1 & 32.6 & 31.6 & 29.8 \\
Post-fusion alignment ($\mathrm{A^{\text{post}}}$) & 38.3 & 35.2& 36.4 & 33.5\\
Prediction-heads & \textbf{56.8} & \textbf{57.1} & \textbf{59.2} & \textbf{56.5}\\
 \bottomrule
\end{tabular}
\caption{Advantage of predicting using prediction heads on decoder. We observe that directly predicting from fusion matrices is inferior as it lacks fine-grained refinement.}
\label{tab:pred_study}
\end{table}

\begin{table}[!t]
\scriptsize
    \centering
\begin{tabular}{c c c c c}
\toprule
 \multicolumn{1}{l}{}  & \multicolumn{2}{c}{\actdata{}}  & \multicolumn{2}{c}{\data{}}\\
 \cmidrule(r){2-3} \cmidrule(r){4-5}
Alignment & mIoU & R@1 & mIoU & R@1\\
 \midrule 
    None  &  50.4 &  51.8 & 52.1 & 48.6\\
    TCC  & 52.4 & 52.7  &  53.9 & 52.8\\
    Drop-DTW & 54.2 & 54.7 & 55.6 & 54.7\\
    soft-DTW  & \textbf{56.8} & \textbf{57.1} & \textbf{59.2}  & \textbf{56.5}\\
 \hline
\end{tabular}
\caption{Ablation on alignment methods.}
\label{tab:align_loss_abl}
\end{table}

\noindent{\textbf{Predictions from Heads vs Alignment Matrices:}}
The results in Table \ref{tab:pred_study} show that prediction from heads outperforms the prediction from both pre-fusion alignment (\(\mathrm{A^{\text{pre}}}\)) and post-fusion alignment (\(\mathrm{A^{\text{post}}}\)) on both datasets. Notably, prediction heads improve mIoU and R@1 by a large margin, demonstrating their effectiveness in accurately retrieving temporal moments. Post-fusion alignment provides better results than pre-fusion alignment. However, the gains are modest compared to the substantial boost from prediction heads. This justifies the need for the decoder.

\noindent{\textbf{Ablation on Alignment Losses:}}
\model{} can incorporate different alignment methods. Table \ref{tab:align_loss_abl} compares the performance of different alignment methods on both \actdata{} and \data{} datasets. Without any alignment (``None''), the model achieves R@1 scores of 51.8\% on \actdata{} and 48.6\% on \data{}. TCC alignment yields an improvement of 0.9\% and 4.2\% on \actdata{} and \data{}, respectively. Drop-DTW is better than TCC. soft-DTW alignment consistently demonstrates superior performance across both datasets, achieving the highest gains of 5.3\% on \actdata{} and 7.9\% on \data{}. 

\noindent{\textbf{Qualitative Results:}}
We present moment retrieval results on a sample from \actdata{} in Fig.~\ref{fig:qual}, where we compare our proposed approach to variant (e) (refer Sec.~\ref{sec:base} for details) of the four best-performing baselines. The video query is a 5-second clip depicting \emph{calf roping}\footnote{A rodeo event where a rider on horseback attempts to catch and tie a calf within a timed competition.}. The target video, with a duration of 24 seconds, showcases this event occurring between 2.1 and 18.1 seconds. Our proposed \model{} model exhibits the highest overlap with the ground truth for the given query, accurately identifying the correct temporal boundaries. Among the baseline methods, CG-DETR and UniVTG achieve strong overlap; however, they are still outperformed by our approach.

\section{Conclusion}
We introduced \model{}, a robust approach for \emph{Vid2VidMR} that combines abstract representation from encoder with fine-grained features from decoder conditioned on aligned query features. It captures semantic and temporal cues for precise moment localization. Our self-supervised pre-training enhances initialization and boosts performance. Extensive experiments on \actdata{} and our new \data{} dataset show that \model{} outperforms strong baselines. Future directions of this work include exploring multimodal queries and developing scalable architectures to enhance broader applicability. 

\noindent\textbf{Acknowledgment:}
This work is supported by the Microsoft Academic Partnership Grant (MAPG) 2023. Yogesh Kumar is supported by a UGC fellowship, Govt. of India.
\bibliographystyle{abbrvnat}
\bibliography{main}
\end{document}